\documentclass{article}
\usepackage{spconf,amsmath,graphicx}
\usepackage{booktabs,amssymb}
\usepackage{graphicx}
\usepackage{multirow}
\usepackage{xcolor}

\usepackage{booktabs,tabularx}

\begin{document}
\title{Towards Robust Partial-Edit Deepfake Localization: An Edit- and Generator-Conditioned Analysis}




\name{
\begin{tabular}{c}
Aishwarya Fursule$^{1}$ \quad
Vamshi Nallaguntla$^{1}$ \quad
Yuan Tseng$^{2}$ \\[2pt]
David Harwath$^{2}$ \quad
Anderson Avila$^{3}$ \quad
Shruti Kshirsagar$^{1}$
\end{tabular}
}

\address{
\begin{tabular}{c}
$^{1}$Wichita State University, USA \qquad
$^{2}$The University of Texas at Austin, USA \\[2pt]
$^{3}$Institut national de la recherche scientifique (INRS-EMT), Canada
\end{tabular}
}

\maketitle
\begin{abstract}
Partial-edit deepfake localization aims to identify manipulated regions
within otherwise authentic speech. Recent systems report strong overall
results, but aggregate scores do not reveal whether this performance
holds across different editing conditions. In this work, we present an
edit- and generator-conditioned analysis of Segment-Aware Localization
(SAL) using WavLM-Large and XLS-R representations on the VoiceCraft and
SSR-Speech subsets of PartialEdit. We categorize edits along three
transcript-derived axes, namely content, meaning change, and editing
operation, and evaluate each generator separately. While both
representations achieve strong overall performance, localization is
noticeably weaker on SSR-Speech. From our experiments, we found that the  deletion is the hardest
operation, and deletions produced by SSR-Speech are the most challenging
condition for both representations. In contrast, edit content, meaning
change, and sentence novelty have smaller effects. These findings show
that a single aggregate score can mask important weaknesses, and that
partial-edit localization should be evaluated separately across
generators and edit types.
\end{abstract}


\section{Introduction}
\label{sec:introduction}

\begin{table}[t]
\centering
\caption{Three-axis transcript taxonomy for PartialEdit.}
\label{tab:taxonomy}
\footnotesize
\setlength{\tabcolsep}{4pt}
\begin{tabularx}{\columnwidth}{@{}l X@{}}
\toprule
\textbf{Category} & \textbf{Example (original $\rightarrow$ edited)} \\
\midrule
\multicolumn{2}{@{}l}{\textbf{Axis 1: Content }} \\
\hline
\textsc{Action}          & He \textit{opened} $\rightarrow$ \textit{closed} the door \\
\textsc{Attribute}       & a \textit{red} car $\rightarrow$ a \textit{blue} car \\
\textsc{Object} & a \textit{cat} $\rightarrow$ a \textit{dog} \\
\textsc{Function}  & \textit{on} the table $\rightarrow$ \textit{under} the table \\
\textsc{Entity}          & \textit{John} $\rightarrow$ \textit{Mike} \\
\textsc{Multi}           & \textit{red car} $\rightarrow$ \textit{blue truck} \\
\textsc{Other}           & \textit{um} $\rightarrow$ \textit{uh} \\
\midrule
\multicolumn{2}{@{}l}{\textbf{Axis 2: Meaning }} \\
\hline
\textsc{Unrelated}       & He is \textit{reading} $\rightarrow$ She is \textit{dancing} \\
\textsc{Opposite}        & He is \textit{happy} $\rightarrow$ He is \textit{sad} \\
\textsc{Related}  & a \textit{cup} $\rightarrow$ a \textit{mug} \\
\textsc{Similar}         & \textit{big} $\rightarrow$ \textit{large} \\
\textsc{Uncertain}       & I \textit{think} $\rightarrow$ I \textit{guess} \\
\textsc{N/A}             & no defined meaning relationship \\
\textsc{Multi}           & multiple meaning changes \\
\midrule
\multicolumn{2}{@{}l}{\textbf{Axis 3: Operation }} \\
\hline
\textsc{Substitution}    & \textit{red} $\rightarrow$ \textit{blue} \\
\textsc{Deletion}        & He is \textit{here} $\rightarrow$ He \ldots \\
\textsc{Insertion}       & He sat $\rightarrow$ He \textit{quickly} sat \\
\textsc{Multi}           & multiple operation types \\
\bottomrule
\end{tabularx}
\end{table}

Recent advances in text-to-speech (TTS) and voice conversion
(VC) have made synthetic speech increasingly realistic,
creating new challenges for reliable audio deepfake detection
\cite{fursule2026you,nallaguntla2026phonemedf}. Most existing
systems classify an entire utterance as real or fake. However,
modern speech-editing systems can replace, insert, or remove
only a short part of an otherwise authentic recording
\cite{rabhi2024audio,zhang2021initial}. In these cases,
utterance-level detection can flag that an audio sample
contains manipulation, but it cannot show where the edited
region occurs. This limitation has driven growing interest in
frame- and segment-level deepfake localization
\cite{zhang2025audio,he2025manipulated}. Recent localization
methods use self-supervised speech representations
\cite{xie2024tdl}, boundary-aware modeling
\cite{zhong2024bam,wu2024cfprf}, temporal-difference learning
\cite{li2025tdam}, and Segment-Aware Localization (SAL)
\cite{mao2026sal}. These methods report strong overall
performance, but a single aggregate score does not reveal whether
a model performs equally well across different kinds of edits.
Strong average performance may still hide a particular generator
or editing operation that the model consistently struggles to
localize.

Recent work has examined broader reliability challenges in audio
deepfake detection, including gender fairness and bias
diagnosis~\cite{fursule2026gender,fursule2026towards}, as well
as generalization across languages, generators, and acoustic
conditions in partial deepfake localization~\cite{tseng2026spade}. Previous studies have examined factors such as splice
discontinuities, real--fake transitions, codec artifacts, and
lexical information
\cite{negroni2024splicing,liu2024how,pimentel2026partial,zhang2026llm}.
However, few studies have examined how localization performance
changes with the nature of the edit itself. Edits differ in the
type of content they change, the semantic relation between the
original and modified text, and the operation that creates the
modification. In addition, the same type of edit may behave
differently when different speech-editing systems produce it.
Examining these conditions separately can therefore reveal
weaknesses that overall evaluation hides. To study this, we conduct a diagnostic analysis on
PartialEdit~\cite{zhang2025partialedit}. We organize edits along
three transcript-derived axes: Content, Meaning, and Operation.
Content describes what type of information the edit modifies,
Meaning describes how the edited text differs semantically from
the original, and Operation describes whether the edit is a
substitution, deletion, insertion, or a combination of operations.
We also retain generator identity, E1 (VoiceCraft) and
E2 (SSR-Speech), to examine whether similar edits show different
localization behavior across generators.
Table~\ref{tab:taxonomy} summarizes the taxonomy.



Our main contributions are:
(1) we propose a three-axis diagnostic framework for examining
localization behavior across content, meaning, operation, and
generator;
(2) we conduct a generator- and edit-conditioned analysis across
WavLM-Large and XLS-R, supported by statistical and worst-group
evaluation; and
(3) we present a sentence-overlap analysis that examines whether
localization performance depends on sentence content seen
during training.

The remainder of this paper is organized as follows.
In Section~\ref{sec:methodology}, we present our methodology,
including the dataset, the transcript-derived edit taxonomy,
the experimental setup, and the evaluation and statistical
analysis procedures. In Section~\ref{sec:results}, we report
overall, generator-wise, and edit-conditioned localization
results, followed by the statistical and sentence-overlap. Finally, in Section~\ref{sec:conclusion},
we summarize our findings and outline directions for future work.

\begin{figure*}[t]
    \centering
    \includegraphics[width=\textwidth]{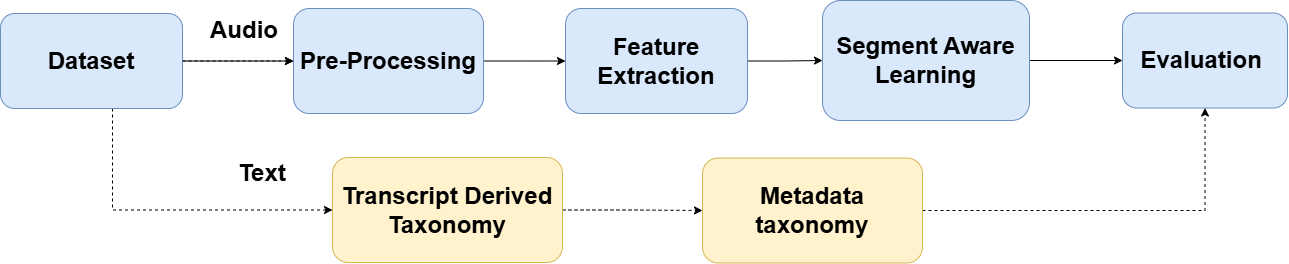}

\caption{Overview of the proposed methodology. The solid path represents the
main audio pipeline used , while the
dotted path represents the offline transcript-derived metadata
pipeline used only for post-hoc evaluation.}

    \label{fig:etr_sal_framework}
\end{figure*}

\section{Proposed Methodology}
\label{sec:methodology}

In this section, we describe the dataset and taxonomy,
audio preprocessing, feature extraction, localization model,
training setup, and evaluation procedure.
\subsection{Dataset and Edit Labels}
\label{subsec:dataset_labels}

We use PartialEdit~\cite{zhang2025partialedit}, a partial speech
deepfake dataset containing speech generated by different editing
systems. We use the publicly available E1 (VoiceCraft) and E2
(SSR-Speech) subsets, since E3 and E4 are not publicly available.
The training, development, and evaluation sets contain 15,704,
15,576, and 53,727 utterances, respectively. For the diagnostic analysis, we compare the original and modified
transcripts and organize each edit along three axes: Content,
Meaning, and Operation. Operation labels are obtained from transcript alignment. Content
categories are derived using spaCy, while Meaning categories use
WordNet and all-MiniLM-L6-v2 sentence similarity. The
Not Applicable (N/A) label is used when a semantic-change
relation is not applicable, while Multi represents edits
involving more than one category or operation. These taxonomy
labels are used only for post-hoc analysis and are never provided
to the localization model during training or inference. All audio is converted to mono 16~kHz before entering the model.
During training, utterances are processed as 4-s segments.
RawBoost is applied with probability 0.5 to introduce additional
acoustic variation. PartialEdit provides genuine/manipulated reference labels at
20-ms resolution. These labels are aligned with the 160-ms
temporal resolution used by the localization model so that each
model output can be compared with the corresponding reference
region.
\subsection{Feature Extraction}
\label{subsec:feature_extraction}

After preprocessing, the waveform is passed to a self-supervised
speech encoder. We use WavLM-Large~\cite{chen2022wavlm} and
XLS-R-300M as two alternative feature extractors. Both models
contain 24 Transformer layers with 16 attention heads and produce
1024-dimensional hidden representations. For both encoders, we use the weighted combination of hidden-layer
representations provided by the S3PRL interface. The resulting
1024-dimensional feature sequence is passed to the same downstream
SAL model. 

\subsection{Segment-Aware Localization Model}
\label{subsec:sal_model}

The extracted feature sequence is passed to the Segment-Aware
Localization (SAL) framework~\cite{mao2026sal}. In our
implementation, both WavLM-Large and XLS-R-300M use the same
SSL\_SeqModel\_8Labels\_2Loss architecture with the
cf sequence model. SAL models the temporal structure of genuine and manipulated
speech segments and produces localization predictions over the
input sequence. Average pooling is used in the downstream model. Cross-Segment Mixing is applied during training to create more
varied segment configurations and reduce dependence on specific
real--fake transition patterns. Importantly, only the audio features are used by SAL. The
transcript-derived Content, Meaning, and Operation labels are
kept separate and are used only during post-hoc evaluation.

\subsection{Experimental Setup}
\label{subsec:experimental_configuration}

Figure~\ref{fig:etr_sal_framework} summarizes the overall methodology.
The solid path shows the audio pipeline used for training and
inference, while the dotted path shows the transcript-derived
metadata used only for post-hoc evaluation.For each speech representation, we train SAL on the combined
E1+E2 training set using the same data splits and training
settings. We use a batch size of 32 and Adam with a learning
rate of $10^{-5}$ and weight decay of $10^{-4}$. Training runs
for up to 50 epochs, with StepLR reducing the learning rate by
a factor of 0.1 every 10 epochs. Cross-Segment Mixing is applied
with a ratio of 0.2 with three seeds: 42, 123, and 2026.
\subsection{Evaluation Metrics}
\label{subsec:evaluation_protocol}

 For evaluation, we report manipulated-class F1 (Fake-F1) and equal error rate
 (EER) as the main metrics at the frame level (160\,ms).
 
The manipulated-class F1 score (Fake-F1) is given by,
\begin{equation}
  \text{Fake-F1} = \frac{2PR}{P+R},
\end{equation}
where $P$ and $R$ denote the precision and recall of frames predicted as
manipulated. Fake-F1 focuses on the edited regions and is not inflated
by the large proportion of genuine frames.

EER is the error rate at the decision threshold where the false
acceptance rate (manipulated frames classified as genuine) equals the
false rejection rate (genuine frames classified as manipulated).
\begin{table}[t]
\centering
\caption{Overall and generator-wise localization (mean $\pm$ standard deviation across three seeds; \%)}
\label{tab:overall_generator}
\footnotesize
\setlength{\tabcolsep}{4pt}
\begin{tabular}{llcc}
\toprule
\textbf{Encoder} & \textbf{Test Set} & \textbf{EER $\downarrow$} & \textbf{Fake-F1 $\uparrow$} \\
\midrule
WavLM-Large & E1+E2 & $2.23\pm0.04$ & $96.15\pm0.08$ \\
XLS-R & E1+E2 & $2.53\pm0.01$ & $95.45\pm0.02$ \\
WavLM-Large & E1 & $0.75\pm0.01$ & $98.86\pm0.01$ \\
XLS-R & E1 & $1.18\pm0.01$ & $97.93\pm0.04$ \\
WavLM-Large & E2 & $3.78\pm0.08$ & $93.20\pm0.16$ \\
XLS-R & E2 & $3.96\pm0.01$ & $92.78\pm0.06$ \\
\bottomrule
\end{tabular}
\end{table}

\begin{table}[t]
\centering
\caption{Overall edit-conditioned performance on PartialEdit
(mean $\pm$ standard deviation over three seeds; \%).}
\label{tab:taxonomy_overall}
\scriptsize
\setlength{\tabcolsep}{2.2pt}
\renewcommand{\arraystretch}{1.05}

\begin{tabular}{clcccc}
\toprule
& &
\multicolumn{2}{c}{\textbf{WavLM-Large}} &
\multicolumn{2}{c}{\textbf{XLS-R}} \\
\cmidrule(lr){3-4}
\cmidrule(lr){5-6}

\rotatebox[origin=c]{90}{\textbf{Axis}} &
\textbf{Category} &
\textbf{EER} &
\textbf{Fake-F1} &
\textbf{EER} &
\textbf{Fake-F1} \\
\midrule

\multirow{7}{*}{\rotatebox[origin=c]{90}{\textbf{Content}}}

& Action
& $1.77\pm0.02$
& $96.96\pm0.07$
& $2.05\pm0.02$
& $96.30\pm0.01$ \\

& Attribute
& $1.86\pm0.05$
& $96.74\pm0.12$
& $2.17\pm0.01$
& $95.98\pm0.04$ \\

& Object
& $2.15\pm0.05$
& $96.14\pm0.08$
& $2.28\pm0.01$
& $95.73\pm0.01$ \\

& Relation
& $2.69\pm0.03$
& $95.20\pm0.10$
& $3.21\pm0.02$
& $94.12\pm0.08$ \\

& Entity
& $2.85\pm0.08$
& $95.36\pm0.09$
& $2.88\pm0.06$
& $95.07\pm0.10$ \\

& Multi
& $3.88\pm0.09$
& $94.10\pm0.11$
& $4.36\pm0.04$
& $93.31\pm0.07$ \\

& Other
& $2.09\pm0.10$
& $94.61\pm0.35$
& $3.91\pm0.09$
& $91.04\pm0.64$ \\

\midrule

\multirow{7}{*}{\rotatebox[origin=c]{90}{\textbf{Meaning}}}

& Unrelated
& $2.10\pm0.04$
& $96.42\pm0.06$
& $2.39\pm0.01$
& $95.76\pm0.03$ \\

& Opposite
& $1.92\pm0.06$
& $96.54\pm0.12$
& $2.24\pm0.03$
& $95.72\pm0.02$ \\

& Related
& $2.01\pm0.03$
& $96.50\pm0.10$
& $2.26\pm0.02$
& $95.82\pm0.03$ \\

& Similar
& $1.86\pm0.06$
& $96.77\pm0.07$
& $2.27\pm0.07$
& $96.06\pm0.36$ \\

& Uncertain
& $2.16\pm0.03$
& $96.20\pm0.06$
& $2.34\pm0.02$
& $95.70\pm0.04$ \\

& N/A
& $4.43\pm0.20$
& $89.79\pm0.49$
& $5.55\pm0.06$
& $87.45\pm0.09$ \\

& Multi
& $3.88\pm0.09$
& $94.10\pm0.11$
& $4.36\pm0.04$
& $93.31\pm0.07$ \\

\midrule

\multirow{4}{*}{\rotatebox[origin=c]{90}{\textbf{Operation}}}

& Substitution
& $2.06\pm0.03$
& $96.42\pm0.07$
& $2.33\pm0.01$
& $95.76\pm0.02$ \\

& Deletion
& $4.81\pm0.37$
& $87.21\pm0.70$
& $6.39\pm0.04$
& $84.01\pm0.19$ \\

& Insertion
& $3.32\pm0.26$
& $94.28\pm0.32$
& $3.70\pm0.11$
& $93.34\pm0.16$ \\

& Multi
& $3.88\pm0.09$
& $94.10\pm0.11$
& $4.36\pm0.04$
& $93.31\pm0.07$ \\
\bottomrule
\end{tabular}
\end{table}

\begin{table*}[t]
\centering
\caption{Generator-specific edit-conditioned performance on PartialEdit
(mean $\pm$ standard deviation over three seeds; \%).}
\label{tab:taxonomy_generators}
\scriptsize
\setlength{\tabcolsep}{2.0pt}
\renewcommand{\arraystretch}{1.05}

\begin{tabular}{clcccccccc}
\toprule
& &
\multicolumn{4}{c}{\textbf{WavLM-Large}} &
\multicolumn{4}{c}{\textbf{XLS-R}} \\
\cmidrule(lr){3-6}
\cmidrule(lr){7-10}

& &
\multicolumn{2}{c}{\textbf{E1}} &
\multicolumn{2}{c}{\textbf{E2}} &
\multicolumn{2}{c}{\textbf{E1}} &
\multicolumn{2}{c}{\textbf{E2}} \\
\cmidrule(lr){3-4}
\cmidrule(lr){5-6}
\cmidrule(lr){7-8}
\cmidrule(lr){9-10}

\rotatebox[origin=c]{90}{\textbf{Axis}} &
\textbf{Category} &
\textbf{EER} &
\textbf{Fake-F1} &
\textbf{EER} &
\textbf{Fake-F1} &
\textbf{EER} &
\textbf{Fake-F1} &
\textbf{EER} &
\textbf{Fake-F1} \\
\midrule

\multirow{7}{*}{\rotatebox[origin=c]{90}{\textbf{Content}}}

& Action
& $0.55\pm0.02$ & $99.15\pm0.02$
& $3.02\pm0.05$ & $94.62\pm0.15$
& $0.95\pm0.01$ & $98.28\pm0.05$
& $3.21\pm0.01$ & $94.19\pm0.06$ \\

& Attribute
& $0.55\pm0.02$ & $99.15\pm0.03$
& $3.19\pm0.09$ & $94.18\pm0.21$
& $0.93\pm0.04$ & $98.26\pm0.05$
& $3.45\pm0.06$ & $93.58\pm0.14$ \\

& Object
& $0.50\pm0.03$ & $99.18\pm0.03$
& $3.85\pm0.10$ & $92.82\pm0.16$
& $0.89\pm0.02$ & $98.37\pm0.03$
& $3.75\pm0.02$ & $92.86\pm0.05$ \\

& Relation
& $0.79\pm0.03$ & $98.72\pm0.03$
& $4.65\pm0.06$ & $91.45\pm0.21$
& $1.37\pm0.05$ & $97.49\pm0.12$
& $5.14\pm0.03$ & $90.57\pm0.05$ \\

& Entity
& $0.68\pm0.08$ & $98.93\pm0.11$
& $5.26\pm0.17$ & $91.08\pm0.11$
& $0.99\pm0.03$ & $98.27\pm0.08$
& $4.98\pm0.10$ & $91.25\pm0.14$ \\

& Multi
& $2.61\pm0.10$ & $96.34\pm0.12$
& $5.26\pm0.09$ & $91.49\pm0.29$
& $3.46\pm0.04$ & $95.12\pm0.08$
& $5.39\pm0.05$ & $91.20\pm0.12$ \\

& Other
& $0.71\pm0.09$ & $98.31\pm0.24$
& $3.42\pm0.41$ & $90.80\pm0.93$
& $1.31\pm0.17$ & $96.90\pm0.27$
& $6.67\pm0.33$ & $85.13\pm0.99$ \\

\midrule

\multirow{7}{*}{\rotatebox[origin=c]{90}{\textbf{Meaning}}}

& Unrelated
& $0.56\pm0.02$ & $99.12\pm0.02$
& $3.67\pm0.09$ & $93.48\pm0.11$
& $1.00\pm0.01$ & $98.23\pm0.05$
& $3.86\pm0.03$ & $93.09\pm0.11$ \\

& Opposite
& $0.60\pm0.00$ & $99.05\pm0.01$
& $3.28\pm0.11$ & $93.88\pm0.24$
& $0.99\pm0.05$ & $98.10\pm0.05$
& $3.54\pm0.02$ & $93.22\pm0.03$ \\

& Related
& $0.61\pm0.01$ & $99.07\pm0.03$
& $3.45\pm0.05$ & $93.74\pm0.18$
& $0.92\pm0.03$ & $98.28\pm0.04$
& $3.67\pm0.03$ & $93.19\pm0.02$ \\

& Similar
& $0.44\pm0.08$ & $99.22\pm0.03$
& $3.32\pm0.07$ & $94.17\pm0.15$
& $1.02\pm0.02$ & $98.26\pm0.11$
& $3.59\pm0.11$ & $93.73\pm0.64$ \\

& Uncertain
& $0.57\pm0.01$ & $99.10\pm0.03$
& $3.82\pm0.03$ & $93.02\pm0.16$
& $1.01\pm0.02$ & $98.17\pm0.02$
& $3.76\pm0.01$ & $93.01\pm0.10$ \\

& N/A
& $1.17\pm0.13$ & $97.29\pm0.26$
& $\mathbf{7.75\pm0.35}$ & $\mathbf{82.62\pm0.78}$
& $2.18\pm0.15$ & $95.22\pm0.42$
& $\mathbf{9.05\pm0.10}$ & $\mathbf{80.16\pm0.47}$ \\

& Multi
& $2.61\pm0.10$ & $96.34\pm0.12$
& $5.26\pm0.09$ & $91.49\pm0.29$
& $3.46\pm0.04$ & $95.12\pm0.08$
& $5.39\pm0.05$ & $91.20\pm0.12$ \\

\midrule

\multirow{4}{*}{\rotatebox[origin=c]{90}{\textbf{Operation}}}

& Substitution
& $0.58\pm0.01$ & $99.10\pm0.02$
& $3.58\pm0.07$ & $93.52\pm0.15$
& $0.99\pm0.02$ & $98.20\pm0.04$
& $3.74\pm0.01$ & $93.13\pm0.07$ \\

& Deletion
& $1.30\pm0.17$ & $96.42\pm0.31$
& $\mathbf{8.27\pm0.63}$ & $\mathbf{79.15\pm1.10}$
& $2.28\pm0.16$ & $94.06\pm0.61$
& $\mathbf{10.38\pm0.15}$ & $\mathbf{75.44\pm0.51}$ \\

& Insertion
& $0.88\pm0.16$ & $98.63\pm0.18$
& $5.97\pm0.68$ & $89.44\pm0.74$
& $1.76\pm0.20$ & $96.96\pm0.15$
& $5.75\pm0.17$ & $89.35\pm0.51$ \\

& Multi
& $2.61\pm0.10$ & $96.34\pm0.12$
& $5.26\pm0.09$ & $91.49\pm0.29$
& $3.46\pm0.04$ & $95.12\pm0.08$
& $5.39\pm0.05$ & $91.20\pm0.12$ \\

\bottomrule
\end{tabular}
\end{table*}

\section{Results and Discussion}
\label{sec:results}

In this section, we present the overall and generator-level localization performance, followed by analyses of the edit taxonomy, statistical consistency, and sentence novelty.

We first evaluate whether localization performance is consistent
across generators. We train models on WavLM-Large and XLS-R
representations with three seeds and evaluate them on the combined
E1+E2 set and on E1 (VoiceCraft) and E2 (SSR-Speech)
separately. Table~\ref{tab:overall_generator} shows that both
representations achieve strong performance on E1+E2, with
WavLM-Large slightly outperforming XLS-R. However, a larger
difference appears when we evaluate the generators separately. On
E2, Fake-F1 decreases by 5.66~pp for WavLM-Large (98.86\% to
93.20\%) and by 5.15~pp for XLS-R (97.93\% to 92.78\%), and EER
increases for both representations. The difference between
generators is therefore several times larger than the difference
between representations. Since all three seeds show the same
trend, the gap does not depend on a particular model
initialization.

Next, we examine which types of edits contribute to localization
difficulty. We group the E1+E2 test utterances along the three
taxonomy axes, namely Content (what the edit changes), Meaning (how
it changes the semantics), and Operation (how it is made), and
compute Fake-F1 and EER for each group.
Table~\ref{tab:taxonomy_overall} shows that the three axes exhibit
different patterns. For Content, both representations perform well
on Action and Attribute edits, while larger drops appear for Multi
and Relation edits; the weakest category differs between
WavLM-Large  and XLS-R. For Meaning, most
categories yield similar results, except N/A, which has the lowest
Fake-F1 for both representations. 
Operation shows the clearest shared trend: both representations
perform best on substitution and worst on deletion, with insertion
and Multi in between. These results indicate that the editing
operation is the axis most consistently related to localization
difficulty.

We then analyze whether operation-level difficulty is uniform
across generators. We repeat the taxonomy analysis separately for
E1 and E2. Table~\ref{tab:taxonomy_generators} shows that
substitution remains the easiest operation for both generators,
whereas deletion shows the largest degradation on E2. Deletion
Fake-F1 decreases from 96.42\% to 79.15\% for WavLM-Large
($-17.27$~pp) and from 94.06\% to 75.44\% for XLS-R
($-18.62$~pp), roughly three times the overall E1--E2 gap. E2
deletion also yields the highest EER among all
generator--operation conditions. This constitutes the main
diagnostic finding of our study: deletion is difficult overall,
and its weakness is substantially stronger for edits produced by
SSR-Speech, consistently across both representations. However,
this comparison does not identify the underlying cause. Edit
duration, acoustic continuity at the edit boundary, the proportion
of manipulated frames, and generator-specific artifacts may all
differ between E1 and E2 deletions. To assess the reliability of the taxonomy effects, we apply a
Kruskal--Wallis test to per-utterance Fake-F1 for each axis,
representation, and seed, report $\epsilon^2$ as the effect size,
and compute the best-to-worst Fake-F1 gap per axis. We also
measure the association between axes using Cram\'er's $V$. All tests
show significant differences ($p<0.05$), but the effect sizes are
small ($\epsilon^2\approx0.008$--$0.010$), indicating that taxonomy
groups explain only a limited part of the per-utterance variation;
given the size of the test set, statistical significance alone is
expected. 

Finally, we investigate whether sentence novelty affects
localization. We split the test set into sentences that appear in
the training data (seen) and sentences that do not (unseen), and
compare their performance. Table~\ref{tab:sentence_overlap_generator}
shows only a small difference between seen and unseen sentences on
E1+E2. At the
generator level, E1 remains nearly unchanged, while E2 is more
sensitive to sentence novelty, particularly for WavLM-Large. These results suggest that
repeated sentence content does not explain the overall
localization behavior, although unseen sentences introduce
additional difficulty for E2.

\begin{table}[t]
\centering
\caption{Generator-wise sentence-overlap performance on PartialEdit
(mean $\pm$ standard deviation over three seeds; \%).}
\label{tab:sentence_overlap_generator}
\scriptsize
\setlength{\tabcolsep}{2.2pt}
\renewcommand{\arraystretch}{1.05}

\begin{tabular}{llcccc}
\toprule
& &
\multicolumn{2}{c}{\textbf{WavLM-Large}} &
\multicolumn{2}{c}{\textbf{XLS-R}} \\
\cmidrule(lr){3-4}
\cmidrule(lr){5-6}

\textbf{Condition} &
\textbf{Gen.} &
\textbf{EER} &
\textbf{Fake-F1} &
\textbf{EER} &
\textbf{Fake-F1} \\
\midrule

\multirow{3}{*}{Seen}
& E1+E2
& $2.08\pm0.04$
& $96.35\pm0.06$
& $2.46\pm0.01$
& $95.45\pm0.04$ \\

& E1
& $0.74\pm0.01$
& $98.88\pm0.02$
& $1.13\pm0.01$
& $97.95\pm0.04$ \\

& E2
& $3.47\pm0.07$
& $93.60\pm0.12$
& $3.85\pm0.04$
& $92.75\pm0.11$ \\

\midrule

\multirow{3}{*}{Unseen}
& E1+E2
& $2.46\pm0.05$
& $95.88\pm0.11$
& $2.63\pm0.02$
& $95.46\pm0.02$ \\

& E1
& $0.77\pm0.02$
& $98.85\pm0.01$
& $1.24\pm0.02$
& $97.90\pm0.03$ \\

& E2
& $4.23\pm0.09$
& $92.65\pm0.21$
& $4.12\pm0.03$
& $92.82\pm0.02$ \\
\bottomrule
\end{tabular}
\end{table}

\section{Conclusion}
\label{sec:conclusion}

In this work, we study partial audio deepfake localization across
generators, edit types, and sentence novelty using WavLM-Large and
XLS-R representations. Both representations achieve strong overall
performance, but this hides a clear weakness: deletion edits
produced by SSR-Speech are the most difficult condition, with
Fake-F1 dropped significantly for both representations.
Sentence novelty and the content and meaning of edits have smaller
effects. In future work, we will control for edit duration and
manipulated-frame proportion to identify the cause of the deletion
gap.

\bibliographystyle{IEEEbib}
\bibliography{strings,refs}

\end{document}